\documentclass[runningheads]{llncs}
\usepackage{eccv}
\usepackage{eccvabbrv}

\usepackage{graphicx}
\usepackage{amsmath}
\usepackage{amssymb}
\usepackage{booktabs}
\usepackage{rotating}
\usepackage{paralist}
\usepackage{pifont}

\usepackage{array}
\usepackage{arydshln}

\usepackage{makecell}

\usepackage{graphicx}
\usepackage{amsmath}
\usepackage{amssymb}
\usepackage{makecell}
\usepackage[accsupp]{axessibility}
\usepackage[accsupp]{axessibility}  
\usepackage[pagebackref,breaklinks,colorlinks,citecolor=eccvblue,linkcolor=eccvblue]{hyperref}
\usepackage{orcidlink}

\usepackage{colortbl} 
\definecolor{LightCyan}{rgb}{0.88,1,1}

\usepackage{multirow}
\usepackage{booktabs}
\usepackage{comment}
\usepackage{color}

\usepackage{textcomp}
\usepackage{siunitx}
\usepackage{tabulary}
\usepackage{makecell}
\usepackage{multirow}
\usepackage{booktabs}
\usepackage{hyperref}
\usepackage{subcaption}

\makeatletter

\newcommand\figref{Figure~\ref}
\newcommand{\tabref}[1]{Table~\ref{#1}}
\newcolumntype{P}[1]{>{\centering\arraybackslash}p{#1}}
\newcolumntype{M}[1]{>{\centering\arraybackslash}m{#1}}
\let\ts@includegraphics\includegraphics

\author{
I-Hsiang Chen$^{1}$,
~~~
Wei-Ting Chen$^{1}$,~~~
Yu-Wei Liu$^{1}$,~~~
Ming-Hsuan Yang$^{2,3}$, ~~~
Sy-Yen Kuo$^{1}$,~~~
\\[0.2cm]
$^1$National Taiwan University~~
$^2$The University of California, Merced~~\\
$^3$Google Research~~
\\[0.2cm]  
}
\institute{}

\usepackage{float}

\usepackage[capitalize]{cleveref}
\crefname{section}{Sec.}{Secs.}
\Crefname{section}{Section}{Sections}
\Crefname{table}{Table}{Tables}
\crefname{table}{Tab.}{Tabs.}

\begin{document}


\title{Improving Point-based Crowd Counting and Localization Based on Auxiliary Point Guidance}

\maketitle


\begin{abstract}
Crowd counting and localization have become increasingly important in computer vision due to their wide-ranging applications.
While point-based strategies have been widely used in crowd counting methods, they face a significant challenge, i.e., the lack of an effective learning strategy to guide the matching process. 
This deficiency leads to instability in matching point proposals to target points, adversely affecting overall performance.
To address this issue, we introduce an effective approach to stabilize the proposal-target matching in point-based methods.
We propose Auxiliary Point Guidance (APG) to provide clear and effective guidance for proposal selection and optimization, addressing the core issue of matching uncertainty.
Additionally, we develop Implicit Feature Interpolation (IFI) to enable adaptive feature extraction in diverse crowd scenarios, further enhancing the model's robustness and accuracy.
Extensive experiments demonstrate the effectiveness of our approach, showing significant improvements in crowd counting and localization performance, particularly under challenging conditions.
The source codes and trained models will be made publicly available.
\end{abstract}

\begin{figure*}[t]
\centering
\includegraphics[width=\linewidth]{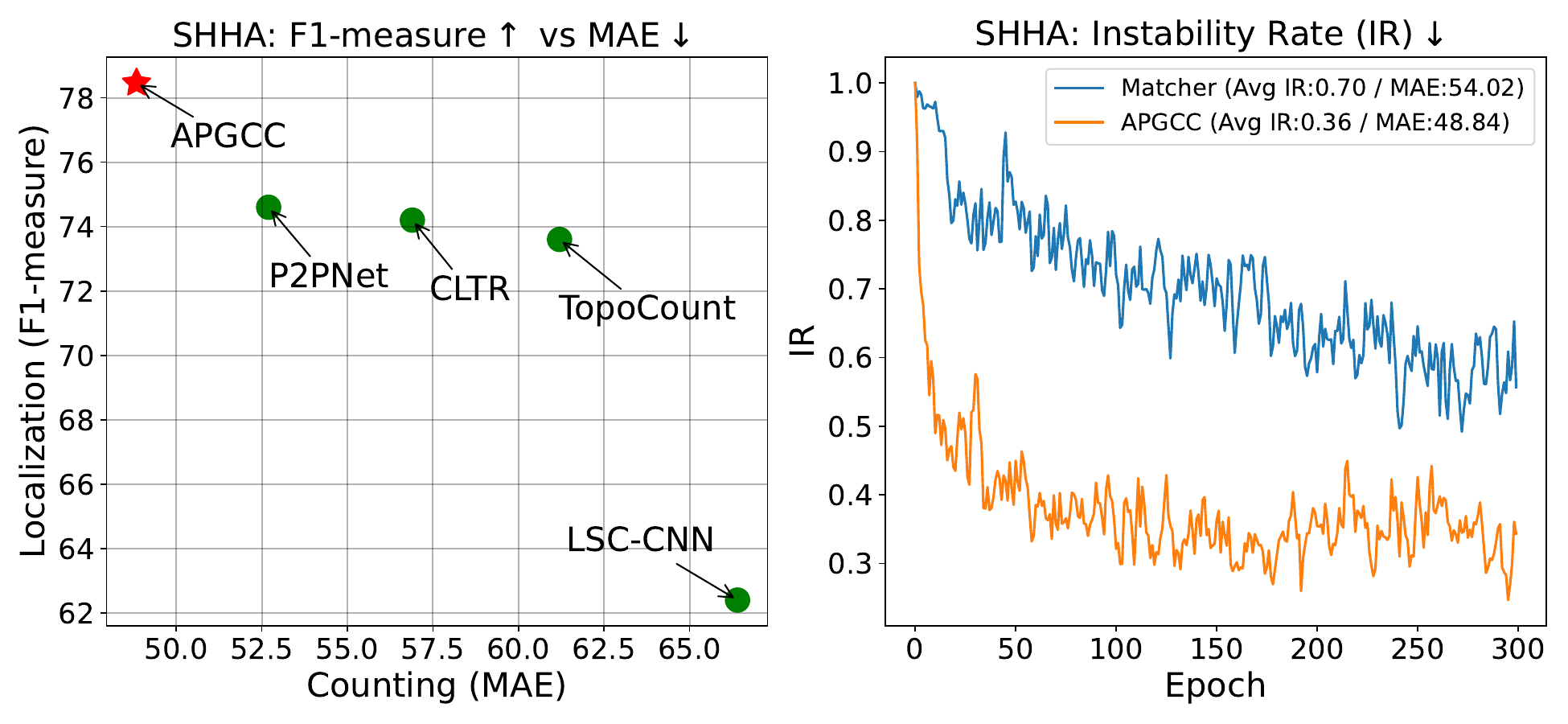}
\label{fig:example2}
\caption{\textbf{(Left) Crowd Counting and Localization:} Comparison with state-of-the-art methods (e.g., LSC-CNN~\cite{sam2020locate}, TopoCount~\cite{abousamra2021localization}, P2PNet~\cite{song2021rethinking} and CLTR ~\cite{liang2022end}) demonstrating the proposed APGCC's effectiveness in accurately counting and localizing in crowded scenes. \textbf{(Right) Matching Process Instability:} Illustrates the instability in selecting point proposals during the matching process by existing point-based methods (e.g., Matcher~\cite{kuhn1955hungarian}) across training epochs, indicated by the Instability Rate (IR), which measures the inconsistency rate of point proposal selection per epoch, leading to limited performance. Both evaluations are conducted on the ShanghaiTech A (SHHA)~\cite{zhang2016single} dataset.}
    \label{fig:examples}
\end{figure*}

\section{Introduction}
Recent years have witnessed the advances and importance of crowd counting and localization in numerous tasks, including surveillance, event management, and urban planning~\cite{jiang2020attention,li2018csrnet,liu2022leveraging,wang2021self,wan2021generalized,gao2019domain,abousamra2021localization,liang2022focal, chen2021all,lin2022boosting,huang2023counting}. 
The pursuit of accurately estimating crowd size and discerning individual locations is fraught with challenges, ranging from fluctuating crowd densities and occlusions to varying environmental settings.

Within the domain of crowd analysis, two principal methodologies emerge: map-based and localization-based approaches. Map-based methods, employing Gaussian kernels to render density maps, effectively provide models with critical information for learning crowd densities. Renowned for their high accuracy in crowd counting, these methods have been validated across a series of studies, including~\cite{jiang2020attention, liu2020adaptive, hu2020count, liu2019counting, liu2020weighing, ma2019bayesian}. 
Despite their capacity to achieve localization through additional designs~\cite{abousamra2021localization,wan2021generalized,idrees2018composition}, they still confront challenges such as the overlapping of maps in densely populated areas and the need for multi-scale representations. This leads to difficulties in precise localization with non-differentiable post-processing techniques like "find-maxima".

Localization-based approaches encompass two divergent strategies: detection-based and point-based methods. Detection-based techniques~\cite{sam2020locate,liu2019point}, characterized by the initiation of pseudo ground truth bounding boxes using nearest-neighbor distances, are tailored for specific scenarios but encounter accuracy limitations in highly congested and sparse areas. Despite their practicality, these methods often contend with the constraints of heuristic post-processing, such as non-maximum suppression, potentially leading to inaccuracies~\cite{liang2022end}.

In contrast, the elegance of point-based methods~\cite{song2021rethinking,liang2022end,liu2023point} lies in directly using point annotations as learning targets. These frameworks can direct the regression of individual coordinates, simplifying the localization process. These methods are celebrated for their simplicity, end-to-end trainability, and independence from complex pre-processing and multi-scale feature map fusion. However, a significant challenge in point-based methods for crowd analysis is the instability of proposal-target matching during training, as illustrated in Figure~\ref{fig:examples}. In each epoch, a large proportion of target points are matched with different point proposals compared to the previous epoch. This issue arises due to the absence of an effective learning strategy to guide the network in consistently selecting the most appropriate proposals during optimization. Consequently, the constantly changing relationships between proposals and targets lead to vague and unclear learning objectives for each proposal. This uncertainty in the learning process often results in localized inaccuracies, manifesting as either underestimation or overestimation in specific areas of crowded scenes.

In this paper, we address the prevailing issue of uncertainty in proposal-target matching within point-based methods for crowd analysis. We introduce a novel learning paradigm, \textbf{A}uxiliary \textbf{P}oint \textbf{G}uidance \textbf{C}rowd \textbf{C}ounting (APGCC), designed to instruct the network on the precise selection and optimization of point proposals for matching with target points. APGCC provides a clear and effective directive, ensuring accurate and informed decisions in the proposal selection and optimization process.

To facilitate the application of APGCC, which necessitates feature extraction at arbitrary positions, we propose a method utilizing Implicit Feature Interpolation. This technique adeptly addresses the challenge of accessing features from diverse locations within the network, thereby ensuring the versatility and efficacy of our model in various crowd scenarios. By enhancing the robustness of the matching relationships between proposals and targets, our approach significantly improves the precision and reliability of crowd analysis models.

Extensive experimental results demonstrate that the proposed APG strategy effectively addresses the instability issues in proposal-target matching during the training process (\textcolor{orange}{orange curve} in Figure~\ref{fig:examples}). 
Moreover, it significantly enhances the performance of crowd counting.
Our method performs robustly and favorably against state-of-the-art schemes in both crowd counting and localization tasks.
We make the following contributions in this work:
\begin{compactitem}
\item We introduce Auxiliary Point Guidance, a novel strategy to address the uncertainty in proposal-target matching within point-based crowd counting methods. APG guides the precise selection and optimization of proposals, enhancing model accuracy.
\item We develop an Implicit Feature Interpolation method, enabling effective feature extraction at arbitrary positions. This technique improves the robustness and versatility of our model, particularly in various crowd scenarios.
\end{compactitem}

\section{Related Work}
In the realm of crowd counting, methodologies are broadly categorized into map-based \cite{li2018csrnet, jiang2020attention, bai2020adaptive, liu2020adaptive, hu2020count,xiong2019open,liu2019counting,liu2020weighing,bai2020adaptive,ma2019bayesian,liu2019counting,miao2020shallow} and localization-based approaches\cite{lian2019density,liu2019point,sam2020locate,liu2019recurrent,laradji2018blobs}, each with distinct strategies and challenges.
\noindent
\smallskip\\
\textbf{Map-based Approaches} use Gaussian kernel density maps to achieve localization and counting. Pioneered by researchers like Idrees~\etal~\cite{idrees2018composition} and Gao~\etal~\cite{gao2019domain}, these methods identify individual positions as peaks on density maps. However, they encounter challenges with overlapping in dense crowds. Innovations such as the Distance Label Map~\cite{xu2022autoscale}, Focal Inverse Distance Transform Map (FIDTM)~\cite{liang2022focal}, and Independent Instance Map (IIM)~\cite{gao2020learning} have been introduced to mitigate these issues, though they still require complex post-processing steps. Concurrently, these approaches also advance crowd counting accuracy by integrating density map values, with enhancements like composition loss~\cite{idrees2018composition} and inter-domain feature segregation~\cite{gao2019domain}. These approaches successfully reduce overlaps in crowded areas, yet they require a post-processing step, such as "find-maxima", to pinpoint individual locations. Additionally, their reliance on multi-scale feature maps adds to their complexity, detracting from their simplicity and elegance.
\noindent
\smallskip\\
\textbf{Localization-based Approaches:} In crowd analysis, localization-based methods integrate both detection-based and point-based strategies. Detection-based approaches, utilizing frameworks like Faster RCNN~\cite{ren2015faster}, focus on generating pseudo bounding boxes using techniques such as nearest neighbor distance, as seen in \cite{sam2020locate}, and a winner-take-all loss for refining box selection, especially beneficial for high-resolution images. Liu~\etal\cite{liu2019point} employs curriculum learning to enhance detection and bounding box prediction accuracy. However, these methods often contend with the challenge of pseudo-ground-truth boxes derived from weak point supervision, which can be particularly unreliable in densely populated areas, thus impeding model training and leading to imprecise box predictions. Additionally, they typically involve Non-Maximum Suppression (NMS) in their box filtering process, which is not designed for end-to-end training~\cite{wang2021self}.

In contrast, point-based approaches like those proposed by Song~\etal\cite{song2021rethinking} (P2PNet), Liang~\etal\cite{liang2022end} (CLTR), and Liu~\etal~\cite{liu2023point} (PET) emphasize directly estimating individual head positions, dynamically adjusting to various crowd densities. These methods significantly enhance the accuracy and process efficiency of localization tasks. Nevertheless, their performance can be limited by the instability of proposal-target matching during training, often leading to regional underestimation or overestimation due to unclear learning objectives for proposals.

\section{Preliminary: Point-based Crowd Counting Framework}
\label{sec:pre}
This framework~\cite{song2021rethinking} comprises three main components essential for point-based crowd counting: Point Proposal Prediction, Proposal-Target Matching, and Loss Calculation.
\noindent\smallskip\\
\textbf{Point Proposal Prediction} involves generating point proposals from the deep feature map $\mathcal{F}_s$ outputted by the backbone network, where $s$ is the downsampling stride, and $\mathcal{F}_s$ has a size of $H \times W$. The process includes two parallel branches: regression for predicting point coordinate offsets and classification for determining confidence scores. Each pixel on $\mathcal{F}_s$ corresponds to a patch in the input image, with a predefined set of reference points $\mathcal{R}={R_k | k \in \{1,...,K}\}$ where $K$ is the total number of reference points. The regression branch outputs $H \times W \times K$ point proposals, with the coordinates of a proposal $\hat{p}_j=(\hat{x}_j, \hat{y}_j)$ calculated as: $\hat{x}_j = x_k + \gamma \Delta_{jx}^k$ and $\hat{y}_j = y_k + \gamma \Delta_{jy}^k$ where $\gamma$ scales the predicted offsets, $\Delta_{jx}^k$ and $\Delta_{jy}^k$ present predicted offsets for its coordinates of a proposal $\hat{p}_j=(\hat{x}_j, \hat{y}_j)$.
\noindent\smallskip\\
\textbf{Proposal-Target Matching} follows the Point Proposal Prediction, utilizing the Hungarian algorithm~\cite{kuhn1955hungarian} as proposal-target matching $
\Omega(\mathcal{P}, \hat{\mathcal{P}}, \mathcal{D})$. This strategy ensures a one-to-one correspondence where each ground truth target from $\mathcal{P}$ is matched with a point proposal in $\hat{\mathcal{P}}$. The matching is based on the pair-wise cost matrix $\mathcal{D}$ of size $N \times M$ where $N$ and $M$ denote the number of ground truth points and point proposals. The matrix $\mathcal{D}$ combines the Euclidean distance between point pairs and the confidence score $\hat{c}_j$ of each proposal, defined as:
\begin{equation}
    \mathcal{D}(\mathcal{P}, \hat{\mathcal{P}}) = \left( \tau \left|\left|p_i - \hat{p}_j\right|\right|_{2} - \hat{c}_j\right)_{i\in N, j\in M},
\end{equation} 
where $\tau$ balances the pixel distance and $\hat{c}_j$ is the confidence score of proposal $\hat{p}_j$.

After the matching process, in the optimal matching results denoted as $\Theta$, each ground truth point $p_i$ is optimally matched to a point proposal $\hat{p}_j$, with the matching result represented by the permutation $\psi = \Theta (\mathcal{P}, \hat{\mathcal{P}}, \mathcal{D})$. 
Thus, $\hat{p}_{\psi(i)}$ is the proposal matched to ground truth point $p_i$. The set of matched proposals, $\hat{\mathcal{P}}{\text{pos}} = {\hat{p}_{\psi(i)} | i\in\{1,...,N\}}$, are considered positives, while the unmatched ones, $\hat{\mathcal{P}}{\text{neg}} = {\hat{p}_{\psi(i)} | i\in\{N+1,...,M\}}$, are negatives.
\noindent\smallskip\\
\textbf{Loss Calculation} integrates Euclidean loss $\mathcal{L}_{loc}$ for point regression and Cross Entropy loss $\mathcal{L}_{cls}$ for proposal classification. The combined loss function $\mathcal{L}_{point}$ is:
\begin{equation}
\mathcal{L}_{cls} = -\frac{1}{M} \left( \sum_{i=1}^N \mathrm{log}\hat{c}_{\psi(i)} + \lambda_1 \sum_{i=N+1}^M \mathrm{log}(1 - \hat{c}_{\psi(i)}) \right),
\end{equation}
\begin{equation}
\mathcal{L}_{loc} = \frac{1}{N} \sum_{i=1}^N \left|\left|p_i - \hat{p}_{\psi(i)}\right|\right|_{2}^2,
\end{equation}
\begin{equation}
\mathcal{L}_{point} = \mathcal{L}_{cls} + \lambda_2 \mathcal{L}_{loc},
\label{eq:ori_loss}
\end{equation}
where $\hat{c}_{\psi(i)}$ is the confidence score of the matched proposal $\hat{p}_{\psi(i)}$, $\lambda_1$ adjusts the impact of negative proposals, and $\lambda_2$ balances the regression loss.

\section{Proposed Method}
While current Point-based Approaches demonstrate promising results in crowd counting and localization, we identified instability in the optimization of the matching process, potentially limiting overall performance. We introduce several components designed to stabilize and enhance the matching mechanism to address this challenge.

\begin{figure*}[t]
\centering \includegraphics[width=0.95\textwidth, page=1]{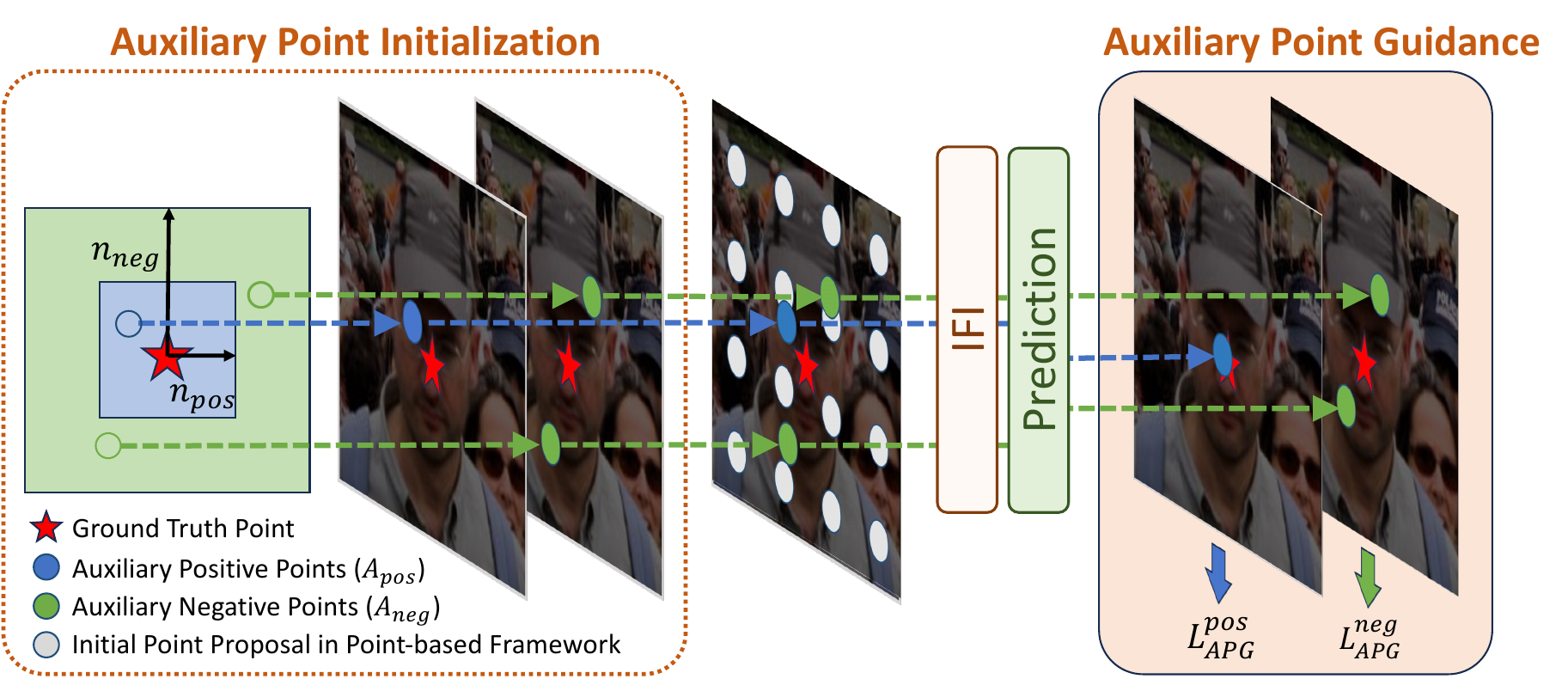}{}
\makeatother 
\caption{\textbf{Illustration of the Auxiliary Point Guidance framework.} During the model's training, we additionally introduce auxiliary positive ($A_{\text{pos}}$) and negative ($A_{\text{pos}}$) points based on each ground truth position to guide the network's learning. This approach helps in directing the optimization process more effectively by distinguishing between potential positive and negative matches.}
\label{fig:apg}
\end{figure*}

\subsection{Auxiliary Point Guidance}
\label{sec:apg}
We introduce an explicit guidance mechanism to enhance the optimization process's stability during the network's matching phase. As shown in \figref{fig:apg}, this involves the strategic designation of auxiliary positive ($A_{\text{pos}}$) and negative ($A_{\text{neg}}$) points within the optimization framework, determined based on ground truth coordinates $(x, y)$. The sets of positive and negative points are defined as $A_{\text{pos}}^{i} = \{(x + R_{\text{pos}}^{i,x}, y + R_{\text{pos}}^{i,y}) \mid i = 1, 2, \ldots, k_{\text{pos}}\}$ and $A_{\text{neg}}^{j} = \{(x + R_{\text{neg}}^{j,x}, y + R_{\text{neg}}^{j,y}) \mid j = 1, 2, \ldots, k_{\text{neg}}\}$. Here, \( R_{\text{pos}}^{i,x} \) and \( R_{\text{pos}}^{i,y} \) represent a series of randomness numbers used to generate the \(x\) and \(y\) coordinates of positive points, respectively, with each number uniformly distributed between \( -n_{\text{pos}} \) and \( n_{\text{pos}} \). Similarly, \( R_{\text{neg}}^{j,x} \) and \( R_{\text{neg}}^{j,y} \) denote series of randomness numbers for generating the \(x\) and \(y\) coordinates of negative points, each uniformly distributed between [\( -n_{\text{neg}} \), \( -n_{\text{pos}} \)] or [\( n_{\text{pos}} \), \( n_{\text{neg}} \)]. The variables \( k_{\text{pos}} \) and \( k_{\text{neg}} \) denote the total number of positive and negative points generated, respectively. Each set \( R_{\text{pos}}^{i} \) and \( R_{\text{neg}}^{j} \) is used to create a unique set of coordinates for \( A_{\text{pos}} \) and \( A_{\text{neg}} \), thereby offsetting the ground truth position \((x, y)\) by these randomness numbers.

Based on the auxiliary positive points $A_{\text{pos}}$, we extract their corresponding features. From these features, we then predict the confidence $\hat{c}_{\text{pos}}^{\star}$ and offset and then calculate the position of the proposal $\hat{p}_{\text{pos}}^{\star}$ for each point. Our objective is to ensure that the confidence of auxiliary positive points is as close to one as possible and that their predicted offsets closely match the added randomness number. To achieve this, we formulate the loss function for the auxiliary positive point as follows:
\begin{equation}
\mathcal{L}^{pos}_{APG} =  \frac{1}{N}\frac{1}{k_{\text{pos}}} \sum_{l=1}^{N} \sum_{i=1}^{k_{\text{pos}}}  (\mathrm{log}\hat{c}_{\text{pos}}^{\star}(l,i) + \lambda_{3}\left ||p_l - \hat{p}_{\text{pos}}^{\star}(l,i)||_{2}^{2}\right) ,
\end{equation}
where $\lambda_{3}$ represents a scaling factor.

For the auxiliary negative points ($A_{\text{neg}}$), our aim is for their confidence $\hat{c}_{\text{neg}}^{\star}$ to be as close to zero as possible. Similarly, we desire their offsets $\Delta_{\text{neg}}^{\star}$ to approach zero, preventing negative points from using offsets to bring their proposal coordinates close to the ground truth. This is crucial to mitigate the potential of these negative points being erroneously considered as matched proposals during the matching process. The loss function specifically formulated for auxiliary negative points is as follows:
\begin{equation}
\mathcal{L}^{neg}_{APG} = \frac{1}{N} \frac{1}{k_{\text{neg}}} \sum_{l=1}^{N} \sum_{j=1}^{k_{\text{neg}}}  (\mathrm{log}(1-\hat{c}_{\text{neg}}^{\star}(l,j)) + \lambda_{4}\left ||\Delta_{\text{neg}}^{\star}(l,j)||_{2}^{2}\right),
\end{equation}
where $\lambda_{4}$ represents a scaling factor.

The total loss of the Auxiliary Point Guidance can be formulated as:
\begin{equation}
\mathcal{L}_{APG}=\mathcal{L}^{pos}_{APG}+\mathcal{L}^{neg}_{APG}.
\label{eq:apg_loss}
\end{equation}
Through this additional guidance, we can direct the network to train point proposals closest to the ground truth points as positive points, while treating those farther away as negative points. This guidance assists the network in consistently selecting the same positive point for each ground truth point during the matching process. Importantly, the chosen positive point is likely to be the correct match, being in close proximity to the ground truth point. By employing this guidance, we address the instability issue inherent in the matching process, thereby enhancing the network's performance.

However, since auxiliary points are randomly assigned based on ground truth coordinates, traditional bilinear interpolation is not suitable for extracting features at these arbitrary positions. Therefore, we propose the use of implicit feature interpolation to obtain these features. The details of this approach will be described in the following section.

\begin{figure*}[t]
\centering \includegraphics[width=0.85\textwidth, page=1]{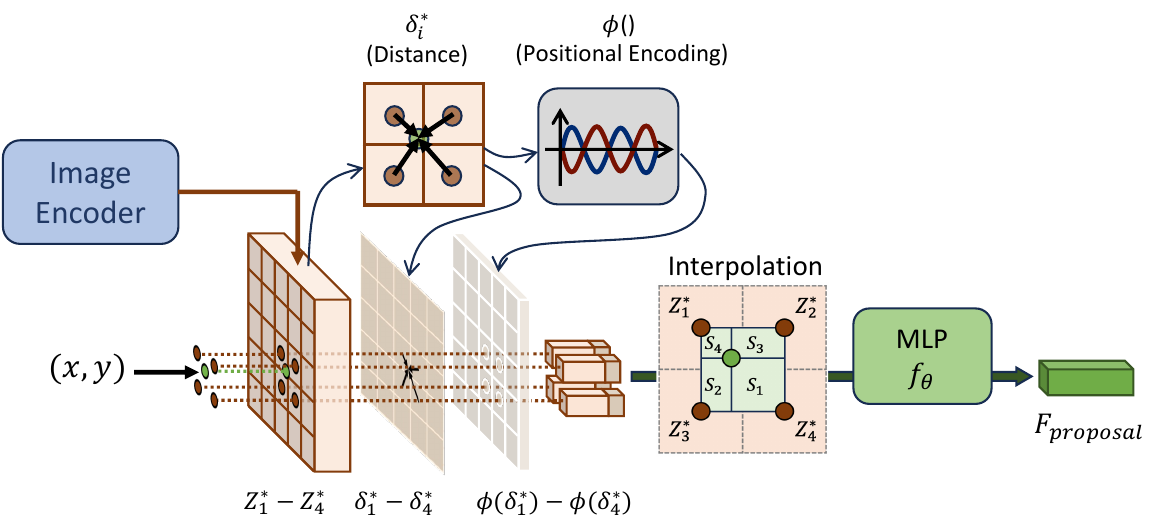}{}
\makeatother 
\caption{\textbf{Illustration of Implicit Feature Interpolation.} Given an arbitrary desired point position $(x, y)$, we concatenate the nearest four feature maps ($Z^{\star}_1$ - $Z^{\star}_4$) along with their distances ($\delta^{\star}_1$ - $\delta^{\star}_4$) to the $(x, y)$ with positional encoding $\phi$ and utilize a Multi-Layer Perceptron (MLP) $f_\theta$ to interpolate the latent feature for that specific location. This approach enables precise feature extraction at non-grid locations, facilitating more flexible and accurate feature representation.}
\label{fig:ifi}
\end{figure*}

\subsection{Implicit Feature Interpolation}
Implicit functions have demonstrated their efficacy in providing robust and continuous feature representations, significantly benefiting various computer vision tasks as evidenced in previous studies~\cite{park2019deepsdf, mildenhall2021nerf, chen2021learning}. In Auxiliary Point Guidance, we leverage implicit function-based interpolation to extract latent features that are both arbitrary and robust. As depicted in \figref{fig:ifi}, for a given point location $(x, y)$, we first determine the four nearest latent features, denoted as $Z^{*}_i| i\in\{1,...,4\}$. We then calculate their respective distances from the target latent feature, represented as $\delta^{*}_i| i\in\{1,...,4\}$. These four latent features, along with their calculated distances, are concatenated channel-wise. This concatenated information is then fed into a MLP to yield the target latent feature. However, it is known that MLPs tend to prioritize low-frequency information, often overlooking crucial high-frequency details, which can impact the performance of the MLP~\cite{basri2020frequency, rahaman2019spectral, tancik2020fourier}. To counter this limitation, we employ positional encoding as suggested in~\cite{xu2021ultrasr}, enhancing the dimensionality of the distance information. By integrating positional encoding with the distance data, we address this high-frequency detail loss. The entire implicit feature interpolation process is encapsulated in the following formulation:
\begin{equation}
F_{proposal}(x,y) = \sum_{i=1}^{4}\frac{S_i}{S}f_{\theta}(Z^{*}_{i},\delta^{*}_{i},\phi(\delta^{*}_{i})),
\end{equation}
where $S_i$ represents the area surrounding the diagonal point with the target point, and $S$ is the sum of these areas, calculated as $S=\sum_{i=1}^{4}S_i$. Here, $f_{\theta}(\cdot)$ symbolizes the MLP, $\phi(\cdot)$ denotes positional encoding, and $F_{proposal}(x,y)$ is the resultant interpolated feature for the point $(x,y)$.

\begin{figure*}[t]
\centering \includegraphics[width=0.95\textwidth, page=1]{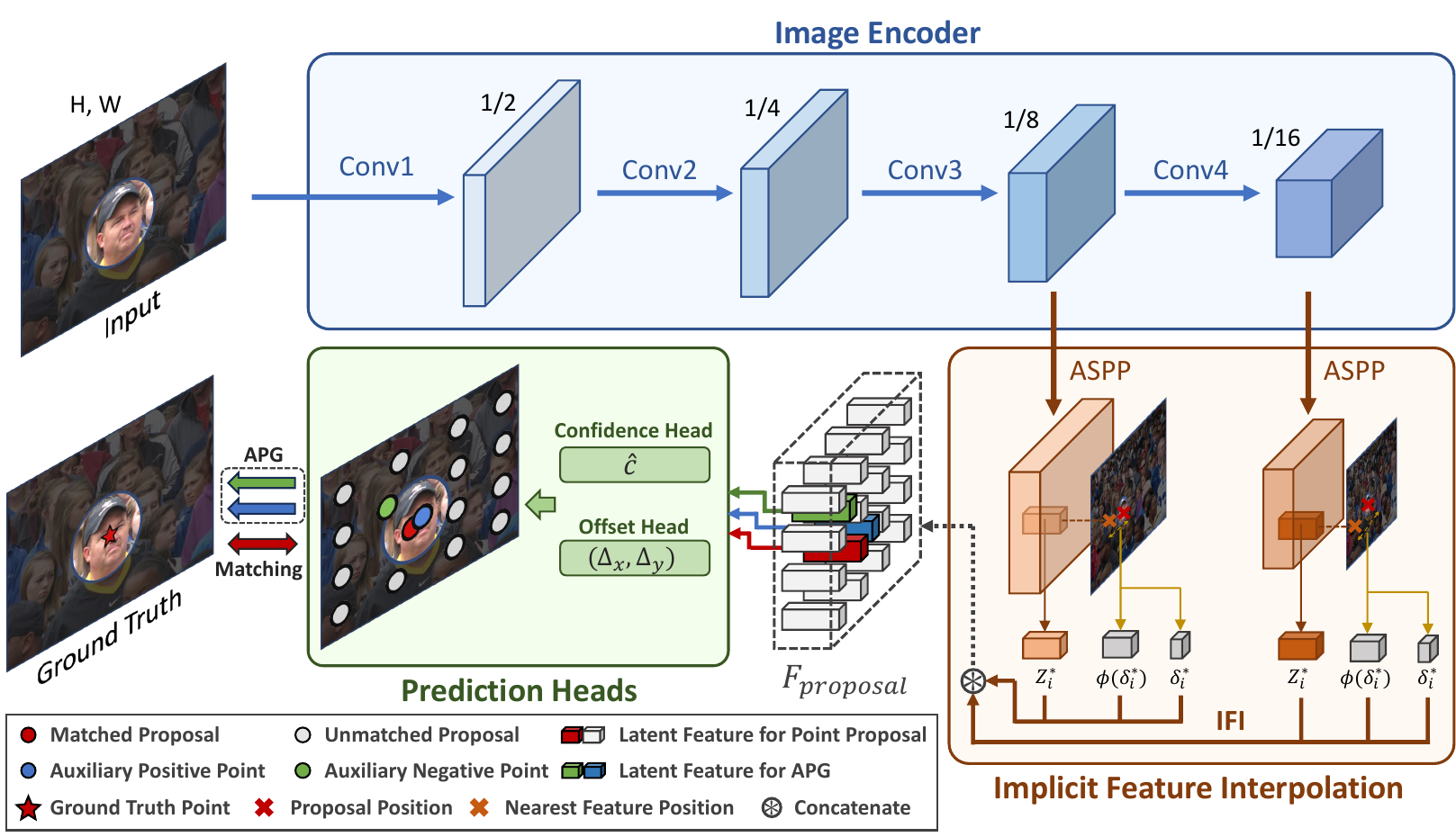}{}
\makeatother 
\caption{\textbf{Illustration of the proposed APGCC for crowd counting and localization.} A VGG encoder extracts image features, where features from conv3 and conv4 layers undergo refinement via Atrous Spatial Pyramid Pooling~\cite{florian2017rethinking}. Subsequently, target latent features are interpolated using implicit feature interpolation. These latent features are then processed through a prediction head to obtain confidence score $\hat{c}$ and offsets $(\Delta_x, \Delta_y)$, facilitating precise crowd counting and localization.}
\label{fig:architecture}
\end{figure*}

\subsection{Architecture Overview}
Our architecture, as illustrated in \figref{fig:architecture}, begins with the extraction of image features using a pre-trained backbone, specifically VGG-16~\cite{simonyan2014very}. We focus on the feature maps from the final two layers (i.e., conv 3 and conv 4). These features are then enhanced for scale diversity through the application of Atrous Spatial Pyramid Pooling (ASPP)~\cite{florian2017rethinking}. Following this, each set of features undergoes a process of implicit feature interpolation, resulting in the computation of corresponding features $F_{proposal}(x,y)$. The interpolated features are subsequently concatenated and input into both the confidence and regression modules. These modules are responsible for predicting the confidence level and offset for each point in the image. The training of the network is accomplished using a combination of the original point-based constraint, as defined in \eqref{eq:ori_loss}, and the proposed Auxiliary Point Guidance, as stated in \eqref{eq:apg_loss}. The total loss function, denoted as $\mathcal{L}_{overall}$, is formulated as follows:
\begin{equation}
\mathcal{L}_{overall}=\mathcal{L}_{point}+\lambda_{5}\mathcal{L}_{APG},
\end{equation}
where $\lambda_{5}$ represents a scaling factor.

\section{Implementation Details}
\subsection{Datasets}
We use the ShanghaiTechA~\cite{zhang2016single}, ShanghaiTechB~\cite{zhang2016single}, UCF\_CC\_50~\cite{idrees2013multi}, UCF-QNRF~\cite{idrees2018composition}, JHU-Crowd++~\cite{sindagi2020jhu}, and NWPU-Crowd~\cite{wang2020nwpu} datasets to evaluate the performance of the proposed method against the state-of-the-art approaches. 
\noindent \smallskip
\textbf{ShanghaiTech A} dataset~\cite{zhang2016single} includes 482 images with 244,167 annotated points. The dataset is divided into 300 training images and 182 testing images.
\noindent \smallskip\\
\textbf{ShanghaiTech B (SHHB)} dataset~\cite{zhang2016single} features 716 images and 88,488 annotated points, with a split of 400 training images and 316 testing images.
\noindent \smallskip\\
\textbf{UCF\_CC\_50} dataset~\cite{idrees2013multi} encompasses 50 images, totaling 63,974 annotated points. We adhere to a five-fold cross-validation as outlined in~\cite{idrees2013multi}.
\noindent \smallskip\\
\textbf{UCF-QNRF} dataset~\cite{idrees2018composition} contains 1,535 high-resolution web-collected images, with over 1.25 million annotated points. It splits into 1,201 training images and 334 testing images, featuring a broad people count range from 49 to 12,865.
\noindent \smallskip\\
\textbf{JHU-Crowd++} dataset~\cite{sindagi2020jhu} comprises 4,372 images, totaling 1.51 million annotated points. It allocates 2,272 images for training, 500 for validation, and reserves 1,600 images for testing.
\noindent \smallskip\\
\textbf{NWPU-Crowd} dataset~\cite{wang2020nwpu} includes 5,109 images with more than 2.13 million annotated points, distributed across 3,109 training images, 500 validation images, and 1,500 testing images.

\subsection{Evaluation Protocol}
\noindent \smallskip
\textbf{Counting Metrics.} We employ Mean Absolute Error (MAE) and Mean Squared Error (MSE) as our primary performance metrics, in line with standard practices in the field, defined as $MAE =\frac{1}{Q}\sum_{i=1}^{Q}|GT_{i}-N_i|$, $MSE =\sqrt{\frac{1}{Q}\sum_{i=1}^{Q}(GT_i-N_i)^2}$, where $Q$ represents the total number of images in the dataset, with $GT_i$ and $N_i$ indicating the actual and predicted crowd counts for the $i$-th image, respectively.

\noindent \smallskip
\textbf{Localization Metrics.} To assess localization accuracy, we utilize Precision (P), Recall (R), and F1-measure (F), following the methodologies in ~\cite{wang2020nwpu, idrees2018composition}. A predicted point is considered a True Positive (TP) if its distance from the corresponding ground truth (GT) point is within a specified threshold $\sigma$. For the NWPU-Crowd dataset \cite{wang2020nwpu}, which includes box-level annotations, $\sigma$ is defined as $\sqrt{(w^2 + h^2)}/2$, where $w$ and $h$ are the width and height of each head. In contrast, for the ShanghaiTech dataset, we apply fixed thresholds of $\sigma = 4$ and $\sigma = 8$.

\subsection{Training Details}
We utilize Adam optimization~\cite{kingma2014adam} with a learning rate of $10^{-4}$ for general model optimization. Given that the VGG-16 backbone network weights are pre-trained on ImageNet, a reduced learning rate of $10^{-5}$ is applied for these components. The initial point proposal stride is set at $s = 8$. The number of reference points $K$ varies depending on the dataset: 4 for most and 8 for the QNRF dataset, aligned with dataset statistics to ensure $M > N$. The prediction head comprises four layers with hidden feature dimensions of [1024, 512, 256, 256], and a shared prediction head is used for our point proposals. For point regression, we set $\gamma$ at 100, and the matching weight term $\tau$ at $5\times10^{-2}$. In auxiliary points learning, the number of positive and negative points (\( k_{\text{pos}} \), \( k_{\text{neg}} \)) are set to (2, 2). Randomness ranges ($n_\text{pos}$, $n_\text{neg}$) are set to (2, 8). The loss coefficients are adjusted as $\lambda_1 = 0.5$, $\lambda_2 = 2\times10^{-4}$, $\lambda_3=2\times10^{-4}$, $\lambda_4=2\times10^{-4}$, and $\lambda_5=0.2$ to balance different term contributions.

Data augmentation involves initial random scaling (factor range: [0.7, 1.3]), ensuring the shorter side is at least 128 pixels. Images are then randomly cropped to 128 × 128 patches and subjected to random flipping with a 0.5 probability. The training batch size is 8. The longer side of each image is restricted to 1920 pixels for UCF-QNRF, JHU-Crowd++, and NWPU-Crowd, while maintaining the original aspect ratio.

\begin{table}[t!]
\centering
\caption{\textbf{Evaluation of crowd counting on SHHA~\cite{zhang2016single}, SHHB~\cite{zhang2016single}, UCF-QNRF~\cite{idrees2018composition}, JHU-Crowd~\cite{sindagi2020jhu} datasets.}}
\label{tab:SOTA1}
\scalebox{0.6}{
\begin{tabular}{ccc|cc|cc|cc|cc} 
\toprule
\multirow{2}{*}{Method} & \multirow{2}{*}{Localization} & \multirow{2}{*}{Manner} & \multicolumn{2}{c}{SHHA} & \multicolumn{2}{c}{SHHB} & \multicolumn{2}{c}{UCF-QNRF} & \multicolumn{2}{c}{JHU-Crowd+}  \\
                        &                   &           & MAE $\downarrow$   & MSE $\downarrow$    & MAE $\downarrow$  & MSE $\downarrow$    & MAE $\downarrow$   & MSE $\downarrow$   & MAE $\downarrow$   & MSE $\downarrow$  \\ 
\hline
\hline
AMSNet~\cite{hu2020count}                       & \ding{55}    & Map-based       & 56.7  & 93.4                           & 6.7  & 10.2                          & 101.8 & 163.2                            & -     & -         \\
SDA+DM~\cite{ma2021towards}                     & \ding{55}    & Map-based       & 55.0  & 92.7                           & -    & -                             & 80.7 & 146.3                             & 59.3 & 248.9      \\
GauNet+CSRNet~\cite{cheng2022rethinking}        & \ding{55}    & Map-based       & 61.2  & 97.8                           & 7.6  & 12.7                          & 84.2  & 152.4                            & 69.4  & 262.4     \\
DC~\cite{xiong2022discrete}                     & \ding{55}    & Map-based       & 61.6  & 96.7                           & 7.1  & 11.1                          & 91.4  & 157.5                            & 67.2  & 288.2     \\ 
ChfL~\cite{shu2022crowd}                        & \ding{55}    & Map-based       & 57.5  & 94.3                           & 6.9  & 11.0                          & 80.3  & \underline{137.6}                & 57.0  & 235.7     \\ 
HMoDE~\cite{du2023redesigning}                  & \ding{55}    & Map-based       & 54.4  & 87.4                           & 6.2  & 9.8                           & -     & -                                & \underline{55.7}  & \textbf{214.6}     \\ 
\hline
LSC-CNN~\cite{sam2020locate}                    & \checkmark   & Detection-based & 66.4  & 117.0                          & 8.1  & 12.7                          & 120.5 & 218.2                            & 112.7 & 454.4     \\
TopoCount~\cite{abousamra2021localization}      & \checkmark   & Detection-based & 61.2  & 104.6                          & 7.8  & 13.7                          & 89.0  & 159.0                            & 60.9  & 267.4     \\
GL~\cite{wan2021generalized}                    & \checkmark   & Map-based       & 61.3  & 95.4                           & 7.3  & 11.7                          & 84.3  & 147.5                            & 59.9  & 259.5     \\
P2PNet~\cite{song2021rethinking}                & \checkmark   & Point-based     & 52.7 & 85.1                          & 6.2 & 9.9                           & 85.3 & 154.5                            & -     & -         \\
CLTR~\cite{liang2022end}                        & \checkmark   & Point-based     & 56.9  & 95.2                           & 6.5  & 10.6                          & 85.8  & 141.3                            & 59.5  & 240.6     \\
PET~\cite{liu2023point}                         & \checkmark   & Point-based     & \underline{49.3} & \underline{78.7}  & \underline{6.1} & \underline{9.6}  & \textbf{79.5}    & 144.3   & 58.5  & 238.0     \\
\textbf{APGCC}                                 & \checkmark   & Point-based     & \textbf{48.8}    & \textbf{76.7}     & \textbf{5.6}    & \textbf{8.7}     & \underline{80.1} & \textbf{136.6}      & \textbf{54.3} & \underline{225.9}    \\
\bottomrule
\end{tabular}}
\end{table}

\begin{figure}[t!]
  \begin{minipage}[t]{1.0\textwidth}
  \begin{minipage}[t]{0.5\textwidth}
\captionof{table}{\textbf{Evaluation of crowd counting on UCF\_CC\_50~\cite{idrees2013multi} dataset.}}
\centering
\label{tab:SOTA2}
        \scalebox{0.6}{
\begin{tabular}{ccccc} 
\toprule
Method  & Localization  & Manner  & MAE $\downarrow$  & MSE $\downarrow$  \\ 
\hline\hline
BL~\cite{ma2019bayesian}                 & \ding{55}  & Map-based   & 229.3         & 308.2         \\
AMSNet~\cite{hu2020count}                & \ding{55}  & Map-based   & 208.4         & 297.3         \\
GauNet+CSRNet~\cite{cheng2022rethinking} & \ding{55}  & Map-based   & 215.4         & 296.4         \\
HMoDE~\cite{du2023redesigning}           & \ding{55}  & Map-based   & \underline{159.6}  & \underline{211.2}         \\
\hline  
P2PNet~\cite{song2021rethinking}         & \checkmark & Point-based  & 172.7        & 256.1         \\
PET~\cite{liu2023point}                  & \checkmark & Point-based  & 159.9        & 223.7         \\
\textbf{APGCC}                           & \checkmark & Point-based  &\textbf{154.8}& \textbf{205.5}\\
\bottomrule
\end{tabular}
}
\end{minipage}
\hspace{0.3em}
 \begin{minipage}[t]{0.45\textwidth}
\captionof{table}{\textbf{Evaluation of crowd counting on NWPU~\cite{wang2020nwpu} dataset.}}
\label{tab:SOTA3}    \centering
\scalebox{0.6}{
\begin{tabular}{ccccc} 
\toprule
Method  & Localization & Manner  & MAE $\downarrow$  & MSE $\downarrow$ \\ 
\hline\hline
NoisyCC~\cite{wan2020modeling}           & \ding{55}  & Map-based  & 96.9           & 534.2                \\
UOT~\cite{ma2021learning}                & \ding{55}  & Map-based  & 87.8           & 387.5                \\
MAN~\cite{lin2022boosting}               & \ding{55}  & Map-based  & 76.5           & 323.0                \\
ChfL~\cite{shu2022crowd}                 & \ding{55}  & Map-based  & 76.8           & 343.0                \\
HMoDE+REL~\cite{du2023redesigning}       & \ding{55}  & Map-based  & \underline{73.4}    & 331.8                \\
\hline
RAZ~\cite{liu2019recurrent}              & \checkmark & Map-based  & 151.4           & 634.6                \\
GL~\cite{wan2021generalized}             & \checkmark & Map-based  & 79.3            & 346.1                \\
AutoScale~\cite{xu2022autoscale}         & \checkmark & Map-based  & 123.9           & 515.5                \\
TopoCount~\cite{abousamra2021localization} & \checkmark & Detection-based & 107.8           & 438.5                \\
P2PNet~\cite{song2021rethinking}         & \checkmark & Point-based & 77.4           & 362.0                \\
CLTR~\cite{liang2022end}                 & \checkmark & Point-based & 74.3            & 333.8                \\
PET~\cite{liu2023point}                  & \checkmark & Point-based & 74.4            & \underline{328.5}    \\
\textbf{APGCC}                           & \checkmark & Point-based &\textbf{71.7}    & \textbf{284.4}       \\
\bottomrule
\end{tabular}
}
    \end{minipage}
\end{minipage}
\end{figure}

\section{Experimental Results}

In this section, we present the evaluation results of the proposed method for crowd counting and localization. 
More results are available in the supplementary material.

\subsection{Evaluation on Crowd Counting}
This section outlines our comparative analysis of crowd counting methods, where our approach is benchmarked against an array of state-of-the-art techniques across diverse datasets. We evaluate our performance against both map-based~\cite{hu2020count, ma2021towards, cheng2022rethinking, xiong2022discrete, shu2022crowd, du2023redesigning, wan2021generalized, ma2019bayesian, wan2020modeling, ma2021learning, lin2022boosting, liu2019recurrent, xu2022autoscale}, detection-based~\cite{sam2020locate, abousamra2021localization} and point-based~\cite{song2021rethinking, liang2022end, liu2023point} methodologies. Our experiments, detailed in Tables~\ref{tab:SOTA1}, ~\ref{tab:SOTA2}, and ~\ref{tab:SOTA3}, highlight APGCC's leading performance, with the \textbf{best} results in bold and the \underline{second-best} results underlined. These findings affirm the effectiveness and adaptability of our approach in various crowd counting scenarios.

Table~\ref{tab:SOTA1} focuses on datasets such as SHHA~\cite{zhang2016single}, SHHB~\cite{zhang2016single}, UCF-QNRF~\cite{idrees2018composition} and JHU-Crowd~\cite{sindagi2020jhu}, showcasing APGCC's significant improvements in accuracy metrics like MAE and MSE. For example, compared to P2PNet~\cite{song2021rethinking} on the SHHA~\cite{zhang2016single} and SHHB~\cite{zhang2016single} datasets, APGCC achieved substantial reductions in both MAE and MSE, demonstrating its effectiveness even in sparse and simple scene conditions.

In Table~\ref{tab:SOTA2}, we specifically examine the UCF\_CC\_50 dataset~\cite{idrees2013multi}, a challenging set of 50 images with complex scenes. Our approach notably excels, achieving impressive results that underscore the efficiency and stability of our learning strategy, particularly beneficial for datasets with limited images.

Finally, Table~\ref{tab:SOTA3} presents our performance on the NWPU-Crowd dataset~\cite{wang2020nwpu}, the most extensive congested dataset considered in our study. Our approach outperforms the competition, including the second-best method, HMoDE+REL~\cite{du2023redesigning}, by achieving lower MAE and MSE. This success can be attributed mainly to our implementation of Auxiliary Points Guidance and the enhancement provided by the Implicit Feature Interpolation technique, which together significantly improve model reliability and adaptability to different scales and densities.

\begin{figure}[t!]
  \begin{minipage}[t]{1.0\textwidth}
  \begin{minipage}[t]{0.5\textwidth}
\captionof{table}{\textbf{Evaluation of crowd localization on NWPU~\cite{wang2020nwpu} dataset.}}
\centering
\label{tab:SOTA4}
        \scalebox{0.55}{
\begin{tabular}{cccccccc} 
\toprule
\multirow{2}{*}{Method} & \multirow{2}{*}{Manner} & \multicolumn{3}{c}{$\sigma_l$ (large threshold)} & \multicolumn{3}{c}{$\sigma_s$ (small threshold)}  \\
\cline{3-8}
                        &         & F(\%) $\uparrow$ & P(\%) $\uparrow$ & R(\%) $\uparrow$     & F(\%) $\uparrow$ & P(\%) $\uparrow$ & R(\%) $\uparrow$             \\ 
\hline
\hline
RAZ~\cite{liu2019recurrent}                     & Map-based    & 59.8   & 66.6   & 54.3              & 57.6   & 47.0   & 51.7               \\
AutoScale~\cite{xu2022autoscale}                & Map-based    & 67.3   & 57.4   & 62.0              & -      & -      & -                  \\
GL~\cite{wan2021generalized}                    & Map-based    & 66.0   & 80.0   & 56.2              & 58.7   & 71.1   & 50.0               \\
\hline        
TinyFaces~\cite{hu2017finding}                  & Detection-based    & 56.7   & 52.9   & 61.1              & 52.6   & 49.1   & 56.6               \\
TopoCount~\cite{abousamra2021localization}      & Detection-based    & 63.7   & 65.1   & 62.4              & -      & -      & -                  \\
\hline        
P2PNet~\cite{song2021rethinking}                & Point-based    & 71.2   & 72.9   & 69.5              & \underline{67.5}      & \underline{68.4}      & \textbf{66.6}      \\
CLTR~\cite{liang2022end}                        & Point-based    & 68.5   & 69.4   & 67.6              & 59.1   & 59.9   & 58.3               \\
PET~\cite{liu2023point}                         & Point-based    & \underline{74.2} & \underline{75.2} & \underline{73.2}  & \underline{67.5}  & \underline{68.4}    & \textbf{66.6}      \\
\textbf{APGCC}                                  & Point-based    & \textbf{76.4}    & \textbf{79.2}    & \textbf{73.6}     & \textbf{68.9}     & \textbf{71.5}       & \underline{66.5}   \\
\bottomrule
\end{tabular}
}
\end{minipage}
\hspace{0.3em}
 \begin{minipage}[t]{0.5\textwidth}
\captionof{table}{\textbf{Evaluation of crowd localization on SHHA~\cite{zhang2016single} dataset.}}
\label{tab:SOTA5}    \centering
\scalebox{0.53}{
\begin{tabular}{cccccccc} 
\toprule
\multirow{2}{*}{Method} & \multirow{2}{*}{Manner} & \multicolumn{3}{c}{$\sigma=4$} & \multicolumn{3}{c}{$\sigma=8$}  \\
\cline{3-8}
                        &                        & F(\%) $\uparrow$ & P(\%) $\uparrow$ & R(\%) $\uparrow$    & F(\%) $\uparrow$ & P(\%) $\uparrow$ & R(\%) $\uparrow$    \\
\hline
\hline
LOBB~\cite{ribera2019locating}               & Map-based         & 25.9   & 34.9   & 20.7      & 53.9   & 67.6   & 44.8       \\
LCFCN~\cite{laradji2018blobs}                & Detection-based   & 32.5   & 43.3   & 26.0      & 56.3   & 75.1   & 45.1       \\
LSC-CNN~\cite{sam2020locate}                 & Detection-based   & 32.6   & 33.4   & 31.9      & 62.4   & 63.9   & 61.0       \\
TopoCount~\cite{abousamra2021localization}   & Detection-based   & 41.1   & 41.7   & 40.6      & 73.6   & 74.6   & 72.7       \\
P2PNet~\cite{song2021rethinking}             & Point-based       & 40.6   & 41.5   & 39.8      & \underline{74.6}   & \underline{76.2}   & 73.1       \\
CLTR~\cite{liang2022end}                     & Point-based       & \underline{43.2}   & \underline{43.6}   & \underline{42.7}      & 74.2   & 74.9   & \underline{73.5}       \\
\textbf{APGCC}                               & Point-based       & \textbf{48.7}   & \textbf{49.2}   & \textbf{48.3}   & \textbf{78.4}   & \textbf{79.1}  & \textbf{77.7} \\
\bottomrule
\end{tabular}}
    \end{minipage}
\end{minipage}
\end{figure}

\subsection{Evaluation on Crowd Localization}
We benchmark our approach against a diverse array of methods, including map-based methods such as RAZ~\cite{liu2019recurrent}, AutoScale~\cite{xu2022autoscale}, LOBB~\cite{ribera2019locating}, and GL~\cite{wan2021generalized}; detection-based methods like TinyFaces~\cite{hu2017finding}, TopoCount~\cite{abousamra2021localization}, LSC-CNN~\cite{sam2020locate}, and LCFCN~\cite{laradji2018blobs}; as well as point-based methods including P2PNet~\cite{song2021rethinking}, CLTR~\cite{liang2022end}, and PET~\cite{liu2023point}.

Table~\ref{tab:SOTA4}, focusing on the NWPU dataset~\cite{wang2020nwpu}, showcases APGCC's superior performance. Compared to detection-based methods that utilize box-level annotations and other point-based approaches, APGCC leverages IFI to acquire precise features and utilizes closer proposal predictions to achieve optimal precision. 
Conversely, in the SHHA dataset~\cite{zhang2016single}, as detailed in Table~\ref{tab:SOTA5}, APGCC secures comprehensive improvements: at a $\sigma=4$, the F1-measure increased by 5.5\%, and at $\sigma=8$, it rose by 3.8\%.

\subsection{Evaluation of Model Complexity}
In Table~\ref{tab:complexity}, we benchmark APGCC against other point-based methods, focusing on the number of parameters and inference time. The inference time evaluations are conducted on an NVIDIA 3090 GPU with an input resolution of $1024\times1024$. The findings illustrate that APGCC maintains efficient computational complexity and delivers superior performance in crowd counting and localization tasks. Note that since our APG training mechanism is employed only during training, it does not incur additional computational overhead during inference. Compared to the original point-based method (i.e., P2PNet~\cite{song2021rethinking}) that utilizes traditional upsampling to process features, our use of IFI allows for more accurate representation learning with fewer parameters. This method enhances computaional efficiency, as employing MLPs for feature interpolation is known to be efficient~\cite{park2019deepsdf}.

\subsection{Ablation Study}
We evaluate the effectiveness of the proposed two modules, APG and IFI. We evaluate the performance on SHHA dataset~\cite{zhang2016single}.
\noindent \smallskip\\
\textbf{Effectiveness of APG.} 
We explore the impact of different optimization strategies, with several distinct settings as follows: (a) "Matcher" solely employs the matching strategy as described in Section \ref{sec:pre} and~\cite{song2021rethinking}, (b) "Nearest Point" directly selects the proposal closest to the ground truth as the positive proposal, with all others considered negative, (c) "APG", which exclusively utilizes the proposed APG for training, and (d) "Matcher + APG" (Ours). The experimental results, as shown in~\tabref{tab:ablation1}, indicate that while strategy (b) may seem intuitive and straightforward to design, it risks multiple ground truths mapping to the same proposal, severely underestimating the final counting. Consequently, using (a) can guide the model on how to allocate proposals for learning and introduce confidence information to enhance discrimination. Our proposed APG effectively addresses the shortcomings of the nearest point, providing an equivalent number of proposals for the model to learn to match the closest proposals. However, as auxiliary positive points cannot be provided during the inference phase, relying solely on APG can make the model overly dependent on reference values. Therefore, by combining the advantages of Matcher and APG (i.e., (d)), we not only teach the model how to allocate a fixed number of proposals but also guide it to make more elegant choices.
\noindent \smallskip\\
\textbf{Optimizing APG Setup.}
Our exploration into optimizing APG focuses on two key aspects: (i) determining the optimal number of potential positive and negative points ($k_\text{pos}$ and $k_\text{neg}$), as defined in Section \ref{sec:apg}, and (ii) adjusting the randomness scale of APG ($n_\text{neg}$ and $n_\text{pos}$), with findings detailed in Tables~\ref{tab:ablation2} and \ref{tab:ablation3}.

Table~\ref{tab:ablation2} explores the impact of the number of auxiliary positive and negative points on performance, essential for validating the APG's effectiveness. The results indicate that while using only auxiliary positive points slightly favors the selection of nearest proposals, it's limited in preventing duplicate predictions. Introducing auxiliary negative points enhances differentiation and training stability by encouraging the model to reject distant proposals. Despite an increase in auxiliary point pairs leading to better stabilization (lower Avg. IR and Avg. $\Delta$), the effect on final performance (MAE) is minimal. Thus, we suggest utilizing a balanced (2, 2) ratio of positive to negative points to achieve better learning performance. This is because training with a (5, 5) setting requires roughly twice the training effort compared to (2, 2), without a significant improvement in performance.

Additionally, the degree of randomness applied to auxiliary proposals plays a pivotal role in the configuration of the APG. Our experiments, executed with a stride of 8, have demonstrated that a precise range of randomness is crucial for attaining optimal results, as evidenced in Table~\ref{tab:ablation3}. While a constrained randomness range may limit the diversity in proposal selection, an excessive range of randomness could jeopardize the model's confidence uniformity across different areas, impacting its effectiveness.
\noindent \smallskip\\
\textbf{Effectiveness of IFI.} 
To accurately capture the correct features for proposals at arbitrary positions, we introduce IFI. Other feasible approaches include: (a) Nearest Neighbor without MLP, which directly utilizes the closest latent feature without any transformation; (b) Bilinear Interpolation without MLP, deriving features at each position through bilinear interpolation, implying no use of coordinate and continuous function transformation; (c) IFI solely employing a single reference point for continuous transformation (MLP with coordinate information); (d) IFI w/o Positional Encoding; and (e) IFI.

The results, as displayed in \tabref{tab:ablation4}, reveal several clear trends. First, the use of interpolation outperforms the nearest-neighbor approach by providing a richer feature context. Second, a comparison between (b) and (d) highlights the benefits of using distance information for continuous transformation. Third, incorporating Positional Encoding significantly aids the MLP in achieving better learning outcomes. By integrating all these methods, we can notably enhance the representation features obtained at any given position.

\begin{figure}[t!]
\centering

\begin{minipage}[b]{0.48\textwidth}
    \centering
    \captionof{table}{\textbf{Comparison of model complexity with Point-based Approaches.}}
    \label{tab:complexity}
    \scalebox{0.65}{
    \begin{tabular}{c|c c c c}
    \toprule
    Method              & P2PNet~\cite{song2021rethinking}  & CLTR~\cite{liang2022end}  & PET~\cite{liu2023point}       & APGCC     \\
    \hline
    Parameters (M) $\downarrow$      & 21.6                            & 43.4                      & \underline{20.9}              & \textbf{18.68}     \\
    Inference Time (s) $\downarrow$     & \underline{0.074}                  & 0.107                     & 0.097            & \textbf{0.071}     \\
    \bottomrule
    \end{tabular}}
\end{minipage}\hfill
\begin{minipage}[b]{0.48\textwidth}
    \centering
    \captionof{table}{\textbf{Analysis on alternatives of optimization strategies.}}
    \label{tab:ablation1} 
    \scalebox{0.8}{
    \begin{tabular}{c|c|cc}
    \toprule
    Setting & Strategy          & MAE $\downarrow$ & MSE $\downarrow$ \\
    \hline
    (a) & Matcher        & 54.04 & 86.97   \\
    (b) & Nearest Point  & 76.91 & 118.60  \\
    (c) & APG            & 58.46 & 96.71   \\
    (d) & Matcher + APG (Ours)  & \textbf{48.84}    & \textbf{76.79}   \\
    \bottomrule
    \end{tabular}}
\end{minipage}

\vspace{\floatsep} 

\begin{minipage}[b]{0.48\textwidth}
    \centering
    \captionof{table}{\textbf{Evaluation of different number of auxiliary points}}
    \label{tab:ablation2} 
    \scalebox{0.55}{
    \begin{tabular}{c|cccccc}
    \toprule
    Num. of Auxiliary Points (\( k_{\text{pos}} \), \( k_{\text{neg}} \)) & (0, 0) & (1, 0)  & (2, 0) & (1, 1) & (2, 2) & (5, 5)    \\
    \hline
    MAE $\downarrow$                                                         & 54.04  & 51.57   & 51.47  & 49.24  & 48.84  & \textbf{48.81} \\
    Avg. IR $\downarrow$                                                     & 0.70   & 0.49    & 0.48   & 0.38   & 0.36   & 0.34  \\
    Avg. $\Delta$ $\downarrow$                                               & 6.87   & 3.89    & 3.37   & 1.62   & 1.58   & 1.49 \\
    \bottomrule
    \end{tabular}}
\end{minipage}\hfill
\begin{minipage}[b]{0.48\textwidth}
    \centering
    \captionof{table}{\textbf{Comparison of different range of randomness for auxiliary points}}
    \label{tab:ablation3} 
    \scalebox{0.65}{
    \begin{tabular}{c|cccc}
    \toprule
    Randomness Range (\( n_{\text{pos}} \), \( n_{\text{neg}} \)) & (1, 4) & (2, 8) & (3, 12)  & (4, 16)   \\
    \hline
    MAE $\downarrow$       & 49.25 & \textbf{48.84} & 50.23 & 51.23  \\
    \bottomrule
    \end{tabular}}
\end{minipage}

\end{figure}

\begin{table}[t!]
\centering
\caption{\textbf{Ablation study of Implicit Feature Interpolation}.}
\label{tab:ablation4}
\scalebox{0.75}{
\begin{tabular}{c|c|cc|cc} 
\toprule
Setting & Method                         & MAE $\downarrow$  & MSE $\downarrow$   & F1@4 $\uparrow$ & F1@8 $\uparrow$             \\
\hline
(a) & Nearest Neighbor without MLP       & 53.16  & 83.31  & 43.59  & 76.27             \\
(b) & Bilinear Interpolation without MLP & 51.25  & 79.92  & 45.78  & 77.67              \\
(c) & IFI with Single Reference Point              & 49.73  & 77.97  & 47.40  & 78.27              \\
(d) & IFI w/o Positional Encoding        & 49.24  & 78.27  & 47.97  & 78.38              \\
(e) & IFI                               & \textbf{48.84}  & \textbf{76.79}  & \textbf{48.76}  & \textbf{78.46}              \\
\bottomrule
\end{tabular}}
\end{table}

\section{Conclusion}
In this paper, we proposed Auxiliary Point Guidance and Implicit Feature Interpolation to address challenges in point-based crowd counting and localization. APGCC improved the stability of proposal-target matching and enabled accurate feature extraction at any position. Extensive experiments against state-of-the-art methods, our approach showed superior performance in various scenarios.


\clearpage





{\small
\bibliographystyle{ieee_fullname}
\bibliography{egbib}

\begin{thebibliography}{10}\itemsep=-1pt

\bibitem{abousamra2021localization}
Shahira Abousamra, Minh Hoai, Dimitris Samaras, and Chao Chen.
\newblock Localization in the crowd with topological constraints.
\newblock In {\em AAAI}, 2021.

\bibitem{bai2020adaptive}
Shuai Bai, Zhiqun He, Yu Qiao, Hanzhe Hu, Wei Wu, and Junjie Yan.
\newblock Adaptive dilated network with self-correction supervision for counting.
\newblock In {\em CVPR}, 2020.

\bibitem{basri2020frequency}
Ronen Basri, Meirav Galun, Amnon Geifman, David Jacobs, Yoni Kasten, and Shira Kritchman.
\newblock Frequency bias in neural networks for input of non-uniform density.
\newblock In {\em ICML}, 2020.

\bibitem{chen2021all}
Wei-Ting Chen, Hao-Yu Fang, Cheng-Lin Hsieh, Cheng-Che Tsai, I Chen, Jian-Jiun Ding, Sy-Yen Kuo, et~al.
\newblock All snow removed: Single image desnowing algorithm using hierarchical dual-tree complex wavelet representation and contradict channel loss.
\newblock In {\em ICCV}, 2021.

\bibitem{chen2021learning}
Yinbo Chen, Sifei Liu, and Xiaolong Wang.
\newblock Learning continuous image representation with local implicit image function.
\newblock In {\em CVPR}, 2021.

\bibitem{cheng2022rethinking}
Zhi-Qi Cheng, Qi Dai, Hong Li, Jingkuan Song, Xiao Wu, and Alexander~G Hauptmann.
\newblock Rethinking spatial invariance of convolutional networks for object counting.
\newblock In {\em CVPR}, 2022.

\bibitem{du2023redesigning}
Zhipeng Du, Miaojing Shi, Jiankang Deng, and Stefanos Zafeiriou.
\newblock Redesigning multi-scale neural network for crowd counting.
\newblock {\em TIP}, 2023.

\bibitem{florian2017rethinking}
L-CCGP Florian and Schroff~Hartwig Adam.
\newblock Rethinking atrous convolution for semantic image segmentation.
\newblock In {\em CVPR}, 2017.

\bibitem{gao2019domain}
Junyu Gao, Tao Han, Qi Wang, and Yuan Yuan.
\newblock Domain-adaptive crowd counting via inter-domain features segregation and gaussian-prior reconstruction.
\newblock {\em arXiv preprint arXiv:1912.03677}, 2019.

\bibitem{gao2020learning}
Junyu Gao, Tao Han, Yuan Yuan, and Qi Wang.
\newblock Learning independent instance maps for crowd localization.
\newblock {\em arXiv preprint arXiv:2012.04164}, 2020.

\bibitem{hu2017finding}
Peiyun Hu and Deva Ramanan.
\newblock Finding tiny faces.
\newblock In {\em Proceedings of the IEEE conference on computer vision and pattern recognition}, pages 951--959, 2017.

\bibitem{hu2020count}
Yutao Hu, Xiaolong Jiang, Xuhui Liu, Baochang Zhang, Jungong Han, Xianbin Cao, and David Doermann.
\newblock Nas-count: Counting-by-density with neural architecture search.
\newblock In {\em ECCV}, 2020.

\bibitem{huang2023counting}
Zhi-Kai Huang, Wei-Ting Chen, Yuan-Chun Chiang, Sy-Yen Kuo, and Ming-Hsuan Yang.
\newblock Counting crowds in bad weather.
\newblock {\em arXiv preprint arXiv:2306.01209}, 2023.

\bibitem{idrees2013multi}
Haroon Idrees, Imran Saleemi, Cody Seibert, and Mubarak Shah.
\newblock Multi-source multi-scale counting in extremely dense crowd images.
\newblock In {\em CVPR}, 2013.

\bibitem{idrees2018composition}
Haroon Idrees, Muhmmad Tayyab, Kishan Athrey, Dong Zhang, Somaya Al-Maadeed, Nasir Rajpoot, and Mubarak Shah.
\newblock Composition loss for counting, density map estimation and localization in dense crowds.
\newblock In {\em ECCV}, 2018.

\bibitem{jiang2020attention}
Xiaoheng Jiang, Li Zhang, Mingliang Xu, Tianzhu Zhang, Pei Lv, Bing Zhou, Xin Yang, and Yanwei Pang.
\newblock Attention scaling for crowd counting.
\newblock In {\em CVPR}, 2020.

\bibitem{kingma2014adam}
Diederik~P Kingma and Jimmy Ba.
\newblock Adam: A method for stochastic optimization.
\newblock {\em arXiv preprint arXiv:1412.6980}, 2014.

\bibitem{kuhn1955hungarian}
Harold~W Kuhn.
\newblock The hungarian method for the assignment problem.
\newblock {\em NRL}, 1955.

\bibitem{laradji2018blobs}
Issam~H Laradji, Negar Rostamzadeh, Pedro~O Pinheiro, David Vazquez, and Mark Schmidt.
\newblock Where are the blobs: Counting by localization with point supervision.
\newblock In {\em ECCV}, 2018.

\bibitem{li2018csrnet}
Yuhong Li, Xiaofan Zhang, and Deming Chen.
\newblock Csrnet: Dilated convolutional neural networks for understanding the highly congested scenes.
\newblock In {\em CVPR}, 2018.

\bibitem{lian2019density}
Dongze Lian, Jing Li, Jia Zheng, Weixin Luo, and Shenghua Gao.
\newblock Density map regression guided detection network for rgb-d crowd counting and localization.
\newblock In {\em CVPR}, 2019.

\bibitem{liang2022end}
Dingkang Liang, Wei Xu, and Xiang Bai.
\newblock An end-to-end transformer model for crowd localization.
\newblock In {\em ECCV}, 2022.

\bibitem{liang2022focal}
Dingkang Liang, Wei Xu, Yingying Zhu, and Yu Zhou.
\newblock Focal inverse distance transform maps for crowd localization.
\newblock {\em TMM}, 2022.

\bibitem{lin2022boosting}
Hui Lin, Zhiheng Ma, Rongrong Ji, Yaowei Wang, and Xiaopeng Hong.
\newblock Boosting crowd counting via multifaceted attention.
\newblock In {\em CVPR}, 2022.

\bibitem{liu2023point}
Chengxin Liu, Hao Lu, Zhiguo Cao, and Tongliang Liu.
\newblock Point-query quadtree for crowd counting, localization, and more.
\newblock In {\em ICCV}, pages 1676--1685, 2023.

\bibitem{liu2019recurrent}
Chenchen Liu, Xinyu Weng, and Yadong Mu.
\newblock Recurrent attentive zooming for joint crowd counting and precise localization.
\newblock In {\em CVPR}, 2019.

\bibitem{liu2019counting}
Liang Liu, Hao Lu, Haipeng Xiong, Ke Xian, Zhiguo Cao, and Chunhua Shen.
\newblock Counting objects by blockwise classification.
\newblock {\em TCSVT}, 2019.

\bibitem{liu2020weighing}
Liang Liu, Hao Lu, Hongwei Zou, Haipeng Xiong, Zhiguo Cao, and Chunhua Shen.
\newblock Weighing counts: Sequential crowd counting by reinforcement learning.
\newblock In {\em ECCV}, 2020.

\bibitem{liu2022leveraging}
Weizhe Liu, Nikita Durasov, and Pascal Fua.
\newblock Leveraging self-supervision for cross-domain crowd counting.
\newblock In {\em CVPR}, 2022.

\bibitem{liu2020adaptive}
Xiyang Liu, Jie Yang, and Wenrui Ding.
\newblock Adaptive mixture regression network with local counting map for crowd counting.
\newblock In {\em ECCV}, 2020.

\bibitem{liu2019point}
Yuting Liu, Miaojing Shi, Qijun Zhao, and Xiaofang Wang.
\newblock Point in, box out: Beyond counting persons in crowds.
\newblock In {\em CVPR}, 2019.

\bibitem{ma2021towards}
Zhiheng Ma, Xiaopeng Hong, Xing Wei, Yunfeng Qiu, and Yihong Gong.
\newblock Towards a universal model for cross-dataset crowd counting.
\newblock In {\em ICCV}, 2021.

\bibitem{ma2019bayesian}
Zhiheng Ma, Xing Wei, Xiaopeng Hong, and Yihong Gong.
\newblock Bayesian loss for crowd count estimation with point supervision.
\newblock In {\em ICCV}, 2019.

\bibitem{ma2021learning}
Zhiheng Ma, Xing Wei, Xiaopeng Hong, Hui Lin, Yunfeng Qiu, and Yihong Gong.
\newblock Learning to count via unbalanced optimal transport.
\newblock In {\em AAAI}, 2021.

\bibitem{miao2020shallow}
Yunqi Miao, Zijia Lin, Guiguang Ding, and Jungong Han.
\newblock Shallow feature based dense attention network for crowd counting.
\newblock In {\em AAAI}, 2020.

\bibitem{mildenhall2021nerf}
Ben Mildenhall, Pratul~P Srinivasan, Matthew Tancik, Jonathan~T Barron, Ravi Ramamoorthi, and Ren Ng.
\newblock Nerf: Representing scenes as neural radiance fields for view synthesis.
\newblock {\em ACM}, 2021.

\bibitem{park2019deepsdf}
Jeong~Joon Park, Peter Florence, Julian Straub, Richard Newcombe, and Steven Lovegrove.
\newblock Deepsdf: Learning continuous signed distance functions for shape representation.
\newblock In {\em CVPR}, 2019.

\bibitem{rahaman2019spectral}
Nasim Rahaman, Aristide Baratin, Devansh Arpit, Felix Draxler, Min Lin, Fred Hamprecht, Yoshua Bengio, and Aaron Courville.
\newblock On the spectral bias of neural networks.
\newblock In {\em ICML}, pages 5301--5310. PMLR, 2019.

\bibitem{ren2015faster}
Shaoqing Ren, Kaiming He, Ross Girshick, and Jian Sun.
\newblock Faster r-cnn: Towards real-time object detection with region proposal networks.
\newblock {\em NeurIPS}, 2015.

\bibitem{ribera2019locating}
Javier Ribera, David Guera, Yuhao Chen, and Edward~J Delp.
\newblock Locating objects without bounding boxes.
\newblock In {\em CVPR}, 2019.

\bibitem{sam2020locate}
Deepak~Babu Sam, Skand~Vishwanath Peri, Mukuntha~Narayanan Sundararaman, Amogh Kamath, and R~Venkatesh Babu.
\newblock Locate, size, and count: accurately resolving people in dense crowds via detection.
\newblock {\em TPAMI}, 2020.

\bibitem{shu2022crowd}
Weibo Shu, Jia Wan, Kay~Chen Tan, Sam Kwong, and Antoni~B Chan.
\newblock Crowd counting in the frequency domain.
\newblock In {\em CVPR}, 2022.

\bibitem{simonyan2014very}
Karen Simonyan and Andrew Zisserman.
\newblock Very deep convolutional networks for large-scale image recognition.
\newblock {\em arXiv preprint arXiv:1409.1556}, 2014.

\bibitem{sindagi2020jhu}
Vishwanath~A Sindagi, Rajeev Yasarla, and Vishal~M Patel.
\newblock Jhu-crowd++: Large-scale crowd counting dataset and a benchmark method.
\newblock {\em TPAMI}, 2020.

\bibitem{song2021rethinking}
Qingyu Song, Changan Wang, Zhengkai Jiang, Yabiao Wang, Ying Tai, Chengjie Wang, Jilin Li, Feiyue Huang, and Yang Wu.
\newblock Rethinking counting and localization in crowds: A purely point-based framework.
\newblock In {\em ICCV}, 2021.

\bibitem{tancik2020fourier}
Matthew Tancik, Pratul Srinivasan, Ben Mildenhall, Sara Fridovich-Keil, Nithin Raghavan, Utkarsh Singhal, Ravi Ramamoorthi, Jonathan Barron, and Ren Ng.
\newblock Fourier features let networks learn high frequency functions in low dimensional domains.
\newblock {\em NeurIPS}, 2020.

\bibitem{wan2020modeling}
Jia Wan and Antoni Chan.
\newblock Modeling noisy annotations for crowd counting.
\newblock {\em NeurIPS}, 2020.

\bibitem{wan2021generalized}
Jia Wan, Ziquan Liu, and Antoni~B Chan.
\newblock A generalized loss function for crowd counting and localization.
\newblock In {\em CVPR}, 2021.

\bibitem{wang2020nwpu}
Qi Wang, Junyu Gao, Wei Lin, and Xuelong Li.
\newblock Nwpu-crowd: A large-scale benchmark for crowd counting and localization.
\newblock {\em TPAMI}, 2020.

\bibitem{wang2021self}
Yi Wang, Junhui Hou, Xinyu Hou, and Lap-Pui Chau.
\newblock A self-training approach for point-supervised object detection and counting in crowds.
\newblock {\em TIP}, 2021.

\bibitem{xiong2019open}
Haipeng Xiong, Hao Lu, Chengxin Liu, Liang Liu, Zhiguo Cao, and Chunhua Shen.
\newblock From open set to closed set: Counting objects by spatial divide-and-conquer.
\newblock In {\em ICCV}, 2019.

\bibitem{xiong2022discrete}
Haipeng Xiong and Angela Yao.
\newblock Discrete-constrained regression for local counting models.
\newblock In {\em ECCV}, 2022.

\bibitem{xu2022autoscale}
Chenfeng Xu, Dingkang Liang, Yongchao Xu, Song Bai, Wei Zhan, Xiang Bai, and Masayoshi Tomizuka.
\newblock Autoscale: learning to scale for crowd counting.
\newblock {\em IJCV}, 2022.

\bibitem{xu2021ultrasr}
Xingqian Xu, Zhangyang Wang, and Humphrey Shi.
\newblock Ultrasr: Spatial encoding is a missing key for implicit image function-based arbitrary-scale super-resolution.
\newblock {\em arXiv preprint arXiv:2103.12716}, 2021.

\bibitem{zhang2016single}
Yingying Zhang, Desen Zhou, Siqin Chen, Shenghua Gao, and Yi Ma.
\newblock Single-image crowd counting via multi-column convolutional neural network.
\newblock In {\em CVPR}, 2016.

\end{thebibliography}
}

\end{document}